\documentclass[journal]{IEEEtran}

\usepackage[hidelinks]{hyperref} 

\usepackage{color,soul}
\sethlcolor{yellow}
\usepackage{xcolor}

\usepackage[none]{hyphenat} 
\usepackage{array}
\newcolumntype{$}{>{\global\let\currentrowstyle\relax}}
\newcolumntype{^}{>{\currentrowstyle}}

\usepackage{booktabs}
\usepackage{multirow}
\usepackage{hyperref}

\usepackage{graphicx} 
\usepackage{float} 
\usepackage{xcolor}
\usepackage{epstopdf}
\usepackage{subfig}
\graphicspath{{figs/}}
\usepackage{tikz}
\usetikzlibrary{shapes.geometric,arrows,matrix,positioning}
\usepackage{pgfplots}

\pgfplotsset{compat=newest} 
\newcounter{plotno}

\usetikzlibrary{pgfplots.groupplots,backgrounds}
\usepgfplotslibrary{units}
\usepackage{tikzscale}

\tikzstyle{io} = [rectangle, minimum width=2cm, minimum height=2cm, text width=2cm, text centered]
\tikzstyle{block} = [rectangle, minimum width=2cm, minimum height=2cm, text width=2cm, text centered, draw=black, line width=0.5mm]
\tikzstyle{arrow_fc} = [line width=1.5mm,->,>=stealth]

\tikzstyle{sum} = [circle, minimum width=0.4cm, draw=black]
\tikzstyle{node1} = [circle, minimum width=0.4cm, draw=black, fill=brown!180]
\tikzstyle{node2} = [circle, minimum width=0.3cm, draw=black, fill=brown!180]
\tikzstyle{node3} = [circle, minimum width=0.2cm, draw=black, fill=brown!180]
\tikzstyle{leaf1} = [rectangle, minimum width=0.2cm, minimum height=0.2cm, draw=black, fill=green]
\tikzstyle{leaf2} = [rectangle, minimum width=0.2cm, minimum height=0.2cm, draw=black, fill=red]
\tikzstyle{arrow} = [line width=0.4mm,-]
\tikzstyle{arrow2} = [line width=0.6mm,->, >=stealth, color=cyan]
\tikzstyle{arrow3} = [line width=0.5mm,->, >=stealth,shorten >= 0.2cm, shorten <= 0.1cm]

\usepackage{mhchem} 
\usepackage{xfrac} 

\usepackage[numbers,sort&compress]{natbib}

\usepackage{setspace}

\usepackage[final]{pdfpages}

\begin{document}

\begin{titlepage}
\setlength{\voffset}{2cm}
This paper is a preprint (IEEE "accepted" status). IEEE copyright notice. \textcopyright~2018 IEEE. Personal use of this material is permitted. Permission from IEEE must be obtained for all other uses, in any current or future media, including reprinting/republishing this material for advertising or promotional purposes, creating new collective works, for resale or redistribution to servers or lists, or reuse of any copyrighted.
\end{titlepage}
\clearpage
\thispagestyle{empty}
\setcounter{page}{1}


%
\title{Machine Learning Cryptanalysis of a~Quantum~Random~Number~Generator}
%
%
%

\author{Nhan~Duy~Truong$^\ast$,~\IEEEmembership{Student Member,~IEEE}, Jing~Yan~Haw$^\ast$, Syed~Muhamad~Assad, Ping~Koy~Lam, Omid~Kavehei,~\IEEEmembership{Senior Member,~IEEE} \thanks{N. D. Truong and O. Kavehei are with \href{www.deepnano.ai}{Nano-Neuro-inspired Research Laboratory}, School of Electrical and Information Engineering, the University of Sydney, Sydney, NSW 2006, Australia.}\thanks{J. Y. Haw, S. M. Assad, P. K. Lam are with Centre for Quantum Computation and Communication Technology, Department of Quantum Science, Research School of Physics and Engineering, the Australian National University, Canberra, ACT 2601, Australia.}\thanks{Correspondence: omid.kavehei@sydney.edu.au.}\thanks{\textit{$^\ast$~N. D. Truong and J. Y. Haw are co-first authors.}}}


%

%
%

\markboth{June~2018}%
{Shell \MakeLowercase{\textit{et al.}}: Bare Demo of IEEEtran.cls for IEEE Journals}
%



\maketitle

\begin{abstract}
Random number generators (RNGs) that are crucial for cryptographic applications have been the subject of adversarial attacks. These attacks exploit environmental information to predict generated random numbers that are supposed to be truly random and unpredictable. Though quantum random number generators (QRNGs) are based on intrinsic indeterministic nature of quantum properties, the presence of classical noise in the measurement process compromises the integrity of a QRNG. In this paper, we develop a predictive machine learning analysis to investigate the impact of deterministic classical noise in different stages of an optical continuous variable QRNG. Our machine learning (ML) model successfully detects inherent correlations when the deterministic noise sources are prominent. After appropriate filtering and randomness extraction processes are introduced, our QRNG system, in turn, demonstrates its robustness against ML. We further demonstrate the robustness of our machine learning approach by applying it to uniformly distributed random numbers from the QRNG and a congruential RNG. Hence, our result shows that machine learning has potentials in benchmarking the quality of RNG devices.
\end{abstract}


%
\IEEEpeerreviewmaketitle

\section{Introduction}
%
%
%
%
\IEEEPARstart{R}{andom} number generators play an important role in cryptographic applications where secure unpredictable keys are necessary~\cite{rukhin2001cryto}. Though pseudo-random number generators (PRNG) can provide uniformly distributed numbers, there is still a risk for security breach because they are based on deterministic algorithms and can exhibit long-range correlation.
For instance, weak cryptographic keys due to the poor source of randomness have been a known threat for years
~\cite{heninger2012WeakKeys,Hastings2016WeakKeys}. In 2012, the biggest scan of Transport Layer Security (TLS) and Secure Shell (SSH) at the time unveiled surprisingly widespread vulnerable keys~\cite{heninger2012WeakKeys}. In fact, the authors in \cite{heninger2012WeakKeys} managed to acquired private keys for $0.5\%$ of HTTPS hosts. Although all major impacted vendors were notified, another survey in 2016 on public internet-wide  TLS keys was still able to reveal over $313,000$ vulnerable keys out of extracted $81$ million distinct Rivest-Shamir-Adleman (RSA) keys~\cite{Hastings2016WeakKeys}. As a result, the National Institute of Standards and Technology (NIST) had to promptly update its recommendation and validation system for deterministic or pseudo-random number generators~\cite{barker2015RNG}.

Meanwhile, since the late 90s, cryptanalytic attacks have been discussed on their applicability to PRNGs \cite{Kelsey1998RNGAtk}. 
In one type of attacks, namely a state compromise extension attack, a PRNG's internal state at some point in time is compromised and used to guess the subsequent outputs \cite{Kelsey1998RNGAtk}. On the other hand, machine learning (ML), which utilizes a large size of training sample to recognize patterns or features in a given dataset, has been an indispensable tool in computer vision, speech recognition and natural language processing \cite{krizhevsky2012imagenet}. Recently, an attack on a weak PRNG, EPC Gen2, was successfully demonstrated in \cite{MeliaSegui2011RNGAtk}. The authors collected pseudo-randomly generated numbers by the linear-feedback shift registers (LFSRs) based EPC Gen2 through eavesdropping communication between an EPC Gen2 and a demo tag to reveal the feedback polynomial function of the LFSRs. The next generated $32$-bit number was predicted with a success rate of $41.5\%$.

To resolve the issue of periodicity, quantum random number generators (QRNG), which are based on intrinsic indeterministic nature of quantum properties are proposed and demonstrated~\cite{kanter2010opticalRNG,herrero2017quantum,ma2016quantum}. This is the implication of Born's rule in quantum mechanics, where the measurement
outcome of a quantum state is inherently probabilistic~\cite{sakurai1995modern}. However, QRNGs should not be assumed to be free from adversarial attacks or fully trusted by default~\mbox{\cite{ma2016quantum,herrero2017quantum}}. These attacks could include monitoring or manipulation of environmental factors, such as temperature, power supply~\mbox{\cite{ma2013postprocessing}} or other types of side-channel attacks~\mbox{\cite{cao2016source,Wei2017QRNGTrust}}. Hence, the effect of an eavesdropper would need to be appropriately considered by means of either device characterization~\mbox{\cite{law2014quantum}}, or other measures based on physical principles such as device-independent~\mbox{\cite{pironio2010random,christensen2013detection}} or semi-device independent~\mbox{\cite{vallone2014quantum,cao2016source}} QRNG to combat against an adversary.

In this paper, we investigate to what extent the classical entropy affects the randomness and unpredictability qualities of a QRNG. In particular, we inspect the classical and the quantum entropy sources in a vacuum fluctuation based QRNG ~\cite{haw2015maximization}. This is done by applying a machine learning based predictive analysis to the hardware processing stages for two scenarios: in the first scenario, we only measure classical noise from the system and in the second scenario, we analyze combined effect of the quantum entropy and classical noise. The robustness of our predictive machine learning analysis not only is demonstrated by outperforming the most probable probability in several instances, but also by successfully predicting the output of a simplified popular PRNG, congruential random number generator with different level of complexity. Our objective is to learn from large raw datasets and environmental data collected from the random number generator and attempt to forecast the next output bit.

\section{Proposed Method}
In this study, we collected large sets of data at multiple stages in both the hardware and software processing stages of the QRNG. 
Inspired by the study in\mbox{~\cite{Kelsey2015PredEntropy}} where the authors suggested the use of machine learning to analyze the previous outputs to guess the next one generated by a PRNG, our aim is to implement a machine learning (ML) tool for predictive data analysis in the absence and presence of the quantum source. Specifically, recurrent convolutional neural network (RCNN) is used to learn potential patterns that may exist among long sequences of generated numbers at different stages of the QRNG. We train the RCNN with $N$ data values to predict the next number to be generated. The probability of a successful output prediction is denoted by $P_{\rm ML}$. This is compared against the guessing probability of the data distribution, i.e., the probability of the most likely outcome, $P_{\rm g}$.

\subsection{QRNG Block Diagram}
\label{sec:qrng_blk}
\begin{figure}[t]
\centering
\includegraphics[width=0.8\columnwidth]{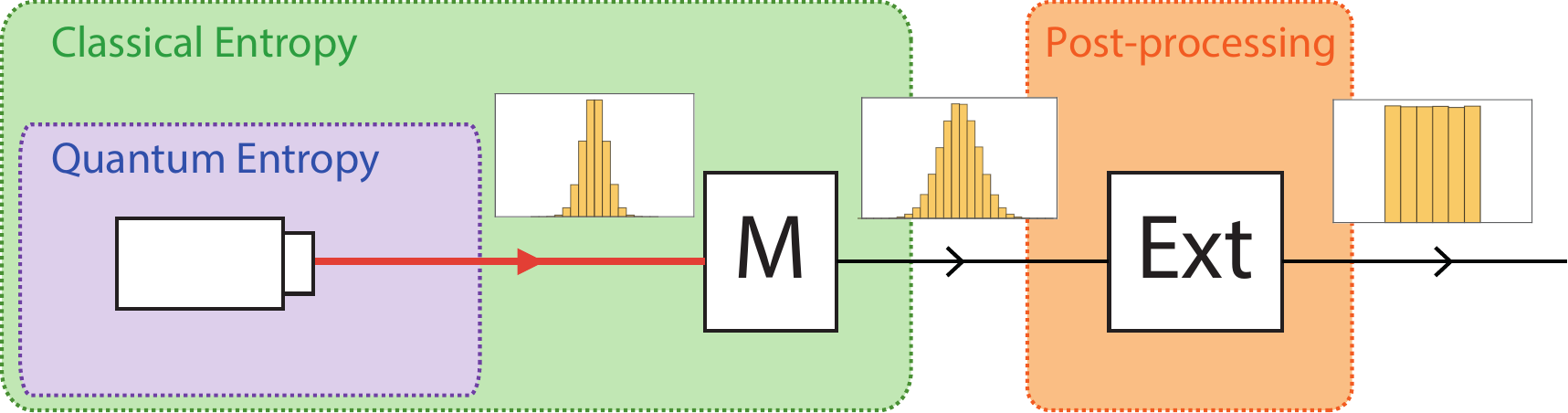}
\caption{Block diagram of the QRNG. A laser source is used to generate quantum entropy. During a measurement (denoted by “M”), statistics of the quantum state is inevitably mixed with the entropy of classical origin. By sacrificing partially random bits, a post-processing randomness extractor stage (denoted by “Ext”) transforms the distribution into a smaller output set with almost uniform distribution.}
\label{fig:QRNG:Qsetup}
\end{figure}

The QRNG can be divided into two segments: entropy source and post-processing procedures~\cite{herrero2017quantum}, as shown in Fig.~\ref{fig:QRNG:Qsetup}. The entropy source produces raw randomness, as a direct result of stochastic physical processes at the source. The raw randomness contains uncertainty of both quantum and classical origin. The classical entropy includes noises from classical devices such as peripheral measuring devices and the analog-to-digital converter (ADC). In order to extract the intrinsic randomness, the classical entropy, which may be untrusted or biased, has to be removed from the main random bit-stream.

The roles of the post-processing block in Fig.~\mbox{\ref{fig:QRNG:Qsetup}} are to extract the quantum randomness out of the physical measurement that combines both quantum and classical randomness and to eliminate biases, hence transforming a non-uniform raw randomness into a supposedly bias-free and unpredictable randomness. The extraction rate depends on the amount of truly unpredictable entropy at the source. 

\subsection{QRNG Setup}
\begin{figure}[b]
\centering
\includegraphics[width=0.95\columnwidth]{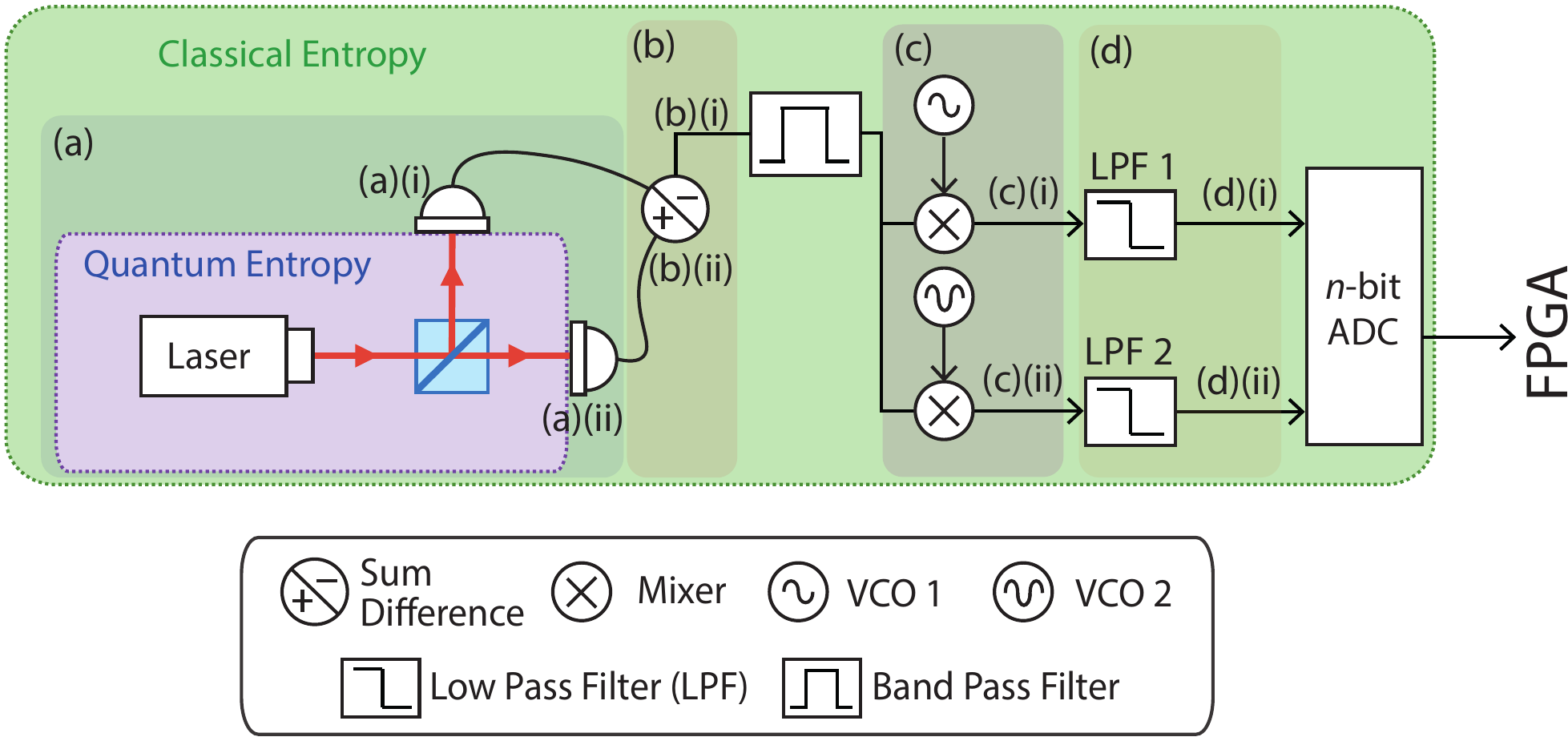}
\caption{Data acquisition stages in the entropy blocks of the QRNG. Stage (a): (i) detector 1 and (ii) detector 2; Stage (b): (i) difference and (ii) sum of the photocurrents; Stage (c): difference of the photocurrents demodulated at (i) $1.375$~GHz (ii)$1.625$~GHz; Stage (d): Low pass filtering of the signals from (c).}
\label{fig:QRNG:Qsetup_extract_data}
\end{figure}

Our source of randomness is based on a continuous variable (CV) quadrature measurement of the vacuum state~\cite{Symul2011,Gabriel2010}. Due to the Heisenberg uncertainty principle, the amplitude quadrature $\hat{X}$ of a vacuum stage is given by the projection of the Wigner function~\cite{Bachor2004}, which follows a Gaussian distribution with zero mean. This distribution can be obtained by performing a homodyne detection, where the vacuum fluctuation is amplified by a strong local oscillator over a balanced beam splitter (Fig.~\ref{fig:QRNG:Qsetup_extract_data}). A homodyne detector can be characterized by its detection bandwidth, the signal-to-noise ratio or dark noise clearance. For our setup, the detection bandwidth is $3$~GHz with about $10$~dB of dark noise clearance.

The output of the two detectors, illustrated as photodetectors in Fig.~\ref{fig:QRNG:Qsetup_extract_data}, are subtracted to eliminate any common-mode noise. In order to avoid technical noise sources at low frequencies, the electronic output from the subtractor is bandpassed ($1$--$3$~GHz) and mixed down at two frequencies: $1.375$~GHz and $1.625$~GHz. The correlations between the sampling points are minimized by passing them through low-pass filters (LPF) with cutoff frequency at $125$~MHz~\cite{shen2010practical}. Finally, these two channels are sampled by $16$-bit ADC at $250$~MSamples per second and sent to field programmable gate array (FPGA) for software processing to extract the quantum randomness out of the measurement outcome.

To investigate the effect of electronic noise within the QRNG setup, data is collected at several stages as indicated in Fig.~\ref{fig:QRNG:Qsetup_extract_data}. These stages are,
\begin{itemize}
\item {\it Stage (a)}\\
(i) Photodetector 1 and (ii) Photodetector 2
\item {\it Stage (b)}\\
(i) Difference and (ii) Sum of the photocurrents from photodetectors 1 and 2
\item {\it Stage (c)}\\
Difference of the photocurrents demodulated at (i) $1.375$~GHz (ii) $1.625$~GHz
\item {\it Stage (d)}\\
Low pass filtered demodulated signal at 
(i) $1.375$~GHz, and (ii) $1.625$~GHz
\end{itemize}
Our experiments are repeated according to two scenarios. First, the classical noise source is probed by removing the quantum source. This can simply be done by turning off local oscillator for the homodyne detection. Second, the convolution of the quantum and classical noise sources are sampled by turning the local oscillator back on. The final post-processed output of the QRNG is also collected for analysis. 

\subsection{Dataset}
At each location, $10$ million continuous data points are captured in the form of $16$-bit integer. The first $5$ million data points are used as training set. The remaining $5$ million is divided into $5$ test-sets, each of which contains $1$ million data points. By having $5$ different test-sets, we can evaluate the consistency in the performance of our deep learning model. Only $13$ most significant bits (MSB) are considered in the learning to reduce memory usage and computational time. Note that these datasets were collected with exactly the same configuration under the two scenarios.

\subsection{Data Preparation}
\begin{figure}[h]
\centering
\includegraphics[width=0.75\columnwidth]{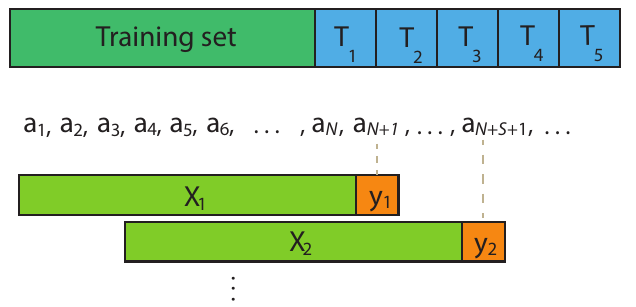}
\caption{Data preparation for each stage. Training set consists of $5$ million samples. Each test-set $\textrm{T}_{k}$ has $1$ million samples. Raw data is a sequence of $13$-bit integers collected at each stage. $N$ neighboring numbers are used as one input sample and the next number is considered as the label.
}
\label{fig:QRNG:QRNG-data}
\end{figure}

We are interested in assessing the possibility of predicting the next number generated by the QRNG at each stage assuming we know a large set of previously generated numbers. Therefore, the raw array of numbers is arranged in such a way that $N$ adjacent numbers are considered as input and the next number is considered as the label (see Fig.~\ref{fig:QRNG:QRNG-data}). Labels are used in our supervised training process to identify our accuracy and other important performance metrics of our deep learning model. In particular, input sample $X_{1}$ is chosen as the first $N$ numbers, while the $(N+1)$-th number is assigned $y_{1}$ as the label for $X_{1}$. The next input sample $X_{2}$ is a shifted version of $X_{1}$ by $S$ positions in the raw array. $S$ is used to control the overlap between samples. In this paper, we use $N=100$ and $S=3$. 

\subsection{Deep Learning}
\label{subsec:DL}
In the recent years, convolutional neural network (CNN) has shown its outstanding capability in recognizing patterns and extracting features in images~\cite{krizhevsky2012imagenet}. Meanwhile, recurrent neural network (RNN), where outputs are dependent on both current inputs and previous computations, is a powerful model in processing sequence data, e.g., language translation. Long-short term memory (LSTM), one of the most commonly used RNN architectures, is shown to be capable of learning long range temporal dependencies and be efficient in addressing sequence to sequence problem~\cite{sutskever2014sequence}. Recurrent convolutional neural network (RCNN), by its name, is a combination of CNN and RNN. RCNNs can be implemented by incorporating recurrent connections into convolutional layers or by feeding features extracted by convolutional layers into an RNN \cite{donahue2015rcnn}.

\begin{figure*}[h]
    \centering
    \includegraphics[width=1.6\columnwidth]{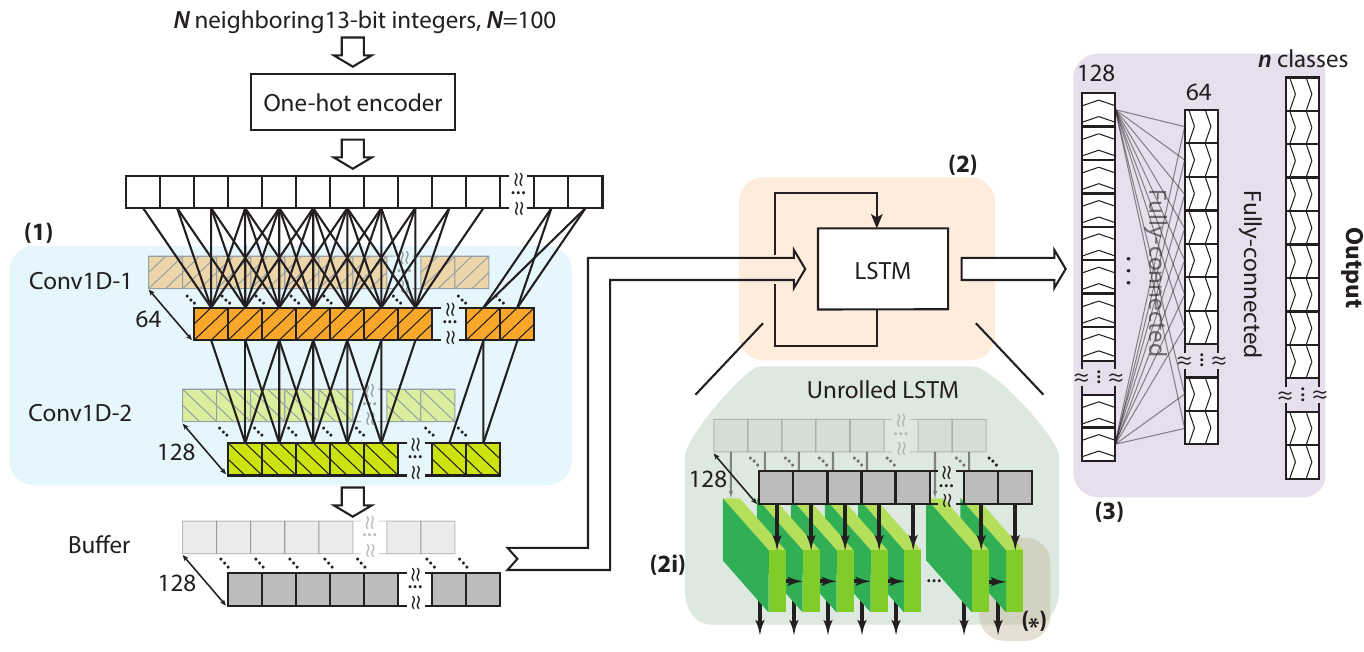}%
    \caption{Recurrent convolutional neural network (RCNN) model: Two convolutional layers, marked by (1), are followed by a LSTM layer, marked by (2), and two fully-connected layers, marked by (3). As a single input into RCNN, $N$~($100$) $13$-bit integers are firstly encoded into one-hot vectors. These $100$ one-hot vectors (blank squares) go to two convolutional layers each of which is followed by a max-pooling of size $2$. The first convolutional layer that has $64$ filters of length $5$, and the second one has $128$ filters with length of $3$. Outputs of the second convolutional layer are fed to the LSTM layer with output size of $128$. An unrolled representation of the LSTM layer is marked by (2i) where all green blocks are LSTM copies at different steps. Each green block takes a piece of input and information from its previous block to generate an output. Output of size $128$ generated the last block (marked by ($\ast$)) that has information of the whole input sequence is connected to $2$ fully-connected layers with output sizes of $64$ and $n$, where $n$ is the number of possible $13$-bit integer values in the dataset.}
    \label{fig:QRNG:QRNG-model}
\end{figure*}

In a QRNG, the electrical noise is ultimately deterministic due to its classical origin, which includes $50$~Hz noise, electronic circuits, electro-magnetic noises, temperature, vibration, and other common-mode signals. Therefore, we utilize LSTM to learn these ``context" using features extracted by convolutional layers. We consider the sequence of generated integers as a text where each integer plays as a word. As illustrated in Fig.~\ref{fig:QRNG:QRNG-model}, $N$~($100$) $13$-bit integers are firstly encoded into one-hot vectors that has all zero elements except a single one element used to distinguish different integer numbers. These $100$ one-hot vectors go to two convolutional layers each of which is followed by a max-pooling of size $2$. The first convolutional layer that has $64$ filters each of which has length of $5$. The second convolutional layer has $128$ filters with length of $3$. Both convolutional layers use rectified linear unit (ReLU) activation functions. Once all outputs of the second convolutional layer are ready, which is guaranteed by a buffer, they are sequentially fed to the LSTM layer configured to have $128$ output size. This operation can be visualized by an unrolled representation of the LSTM layer which is marked by (2i) in Fig.~\ref{fig:QRNG:QRNG-model}. Green blocks are LSTM copies at different steps. Each green block takes a piece of input and information from its previous block to generate an output. We are interested in the output generated by the last block as it has information of the whole sequence. The LSTM output of size $128$ is connected to two fully-connected layers, i.e., each output of previous layer is connected to all input of current layer, with sigmoid and softmax functions, respectively, as activation functions. The two fully-connected layers have output sizes of $64$ and $n$, where $n$ is the number of possible $13$-bit integer values in the dataset.

\subsection{System evaluation}
\begin{figure}[t]
\centering
\includegraphics[width=0.9\columnwidth]{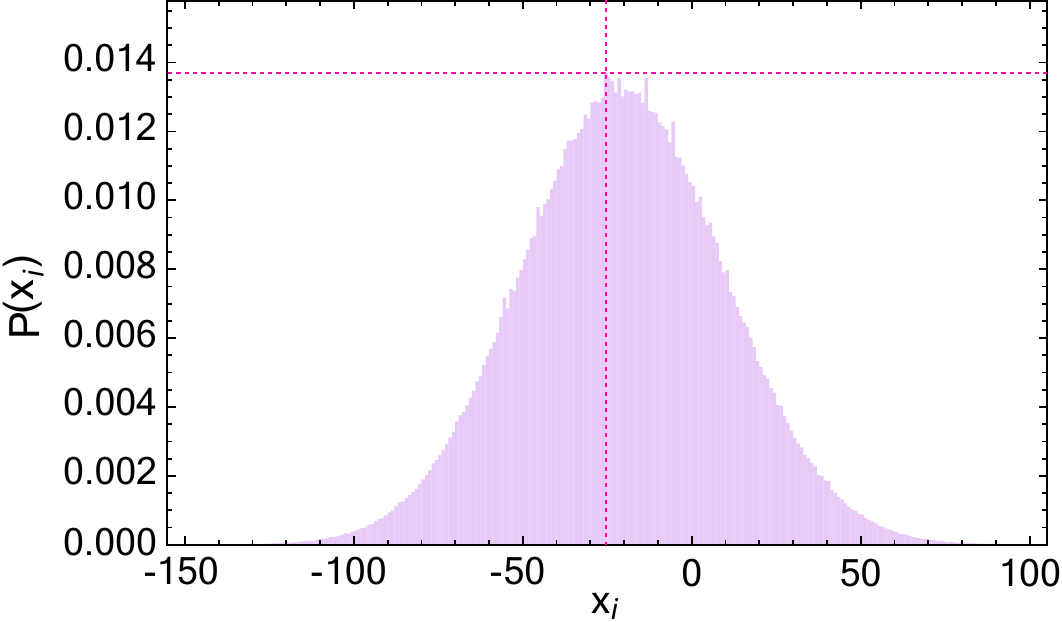}
\caption{Probability distribution of the quadrature measurement of the vacuum. Maximum $P_X(x_i)$ of this set is $0.0137$ when $x_i=-26$. 
}
\label{fig:QRNG:maxbin}
\end{figure}

The performance of our machine learning based prediction model is evaluated by comparing the success probability of the prediction, $P_{\rm ML}$, against the guessing probability of the distribution observed on the QRNG output, $P_{\rm g}$. The successful prediction probability $P_{\rm ML}$ is defined as follows: for a machine that used a training sequence of length $K$ to conduct the learning procedure, $P_{\rm ML} $ is the probability of guessing the $(N+1)$-th number correctly, conditioned on knowing the previous $N$ numbers. On the other hand, the guessing probability $P_{\rm g}$ emerges naturally as the figure of merit by defining randomness as the unpredictability of a given distribution on a single use. This quantity tells us what is the best chance we have in predicting the outcome of a random variable $X$ from a distribution $P_X(x_i)$. In the unit of bits, this quantity is linked to the min-entropy~\cite{Konig2009}, which is defined as:
\begin{align}
H_{\rm min}(X)&=-\log_2\left[P_{\rm g}\right]\nonumber\\
&=-\log_2\left[\max_{x_i \in X} P_X(x_i)\right].
\label{eq:Hmindef}
\end{align} 
Operationally, this quantity tells us the peak of (almost) uniform randomness that can be extracted out of the distribution $P_X(x_i)$. In a given dataset, $P_{\rm g}$ corresponds to the max unit bin guessing of a set of integers, which is the highest probability of a single integer in the distribution of the set $\max_{x_i \in X} P_X(x_i)$. For example, Fig.~\ref{fig:QRNG:maxbin} displays a typical distribution of integers from homodyne detection of the vacuum state. 
With the knowledge of the distribution, the best strategy for an eavesdropper would be to guess the value of $-26$, giving a success rate of $1.37\%$. In our experiment, the max unit bin guessing is learned from training sets and compared to prediction accuracy of the machine learning model. If there exists inherent patterns or predictability among outputs at a certain stage of the QRNG in Fig.~\ref{fig:QRNG:Qsetup_extract_data}, the prediction systems will give a better accuracy compared to the max unit bin guessing.

\section{Results}
\begin{figure}[t]
\centering
\subfloat{
\put(26,296){(a)}
\put(26,130){(b)}
	\includegraphics[width=1\columnwidth]{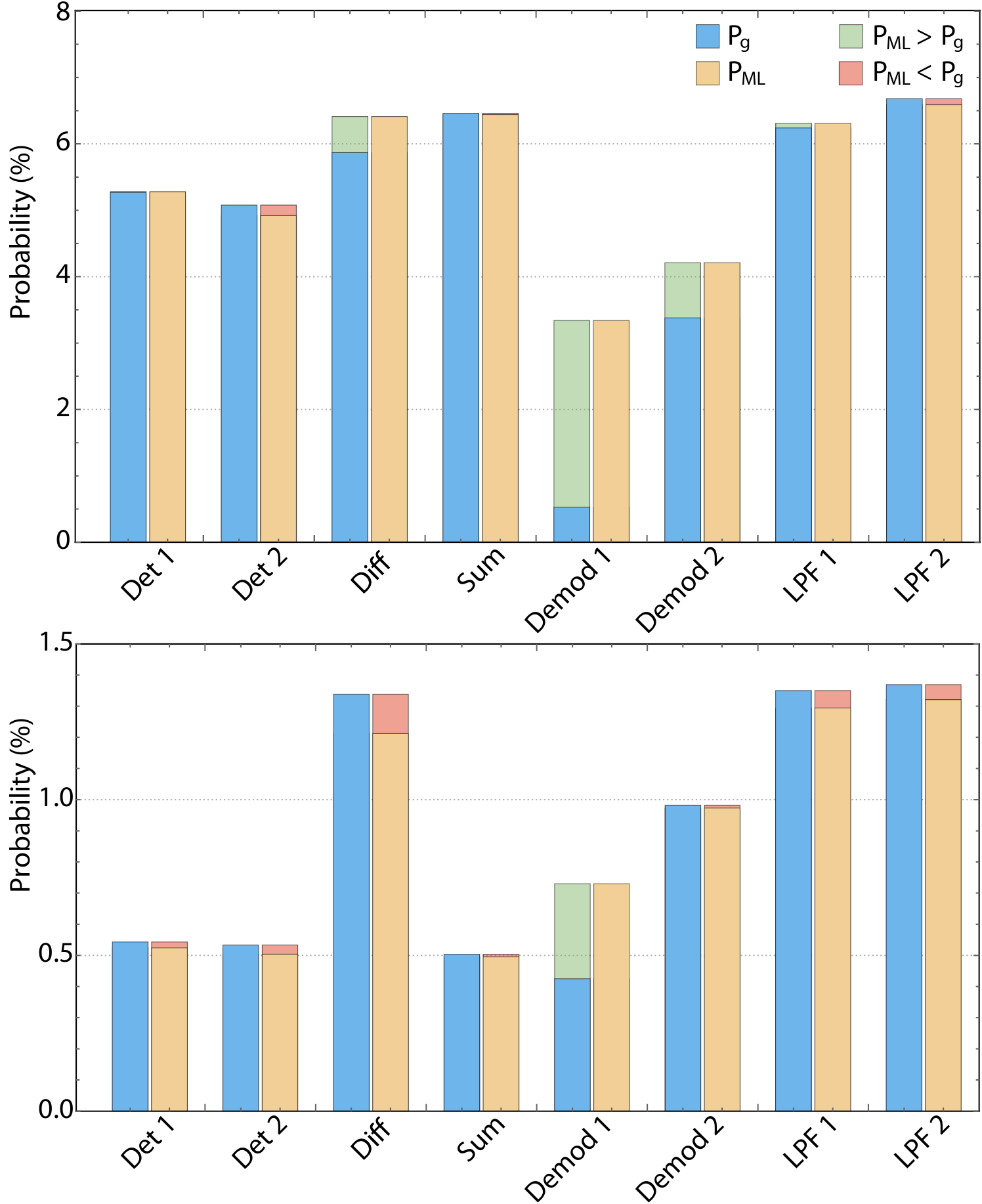}}%

\caption{Prediction performance of the deep learning model. (a) $P_{\rm ML}$ surpasses $P_{\rm g}$ in $4$ out of $8$ stages (Diff, Demod 1, Demod 2 and LPF 1) in scenario 1 (classical) by more than $2$ standard deviations (not shown on the plots) and (b) only $1$ out of $8$ stages (Demod 1) in scenario two (quantum and classical). 
}
\label{fig:QRNG:performance}
\end{figure}

The deep learning model was trained with $5$ million $13$-bit numbers. During the training, $20\%$ of the data were used as validation to monitor whether the model starts to overfit the training set. The maximum number of epochs was set at $20$ and the training would be stopped if validation error ceases to improve after $4$ consecutive epochs. Validation error and corresponding trained weights are recorded during the training. We chose the trained weights with the least validation error as the final trained weights. The five $1$-million number datasets are used to evaluate the model. Fig.~\ref{fig:QRNG:performance} demonstrates the performance of the two models in comparison with a max bin probability for each dataset. 

For the first scenario, where only noises of classical origin are considered, RCNN model shows its capability in learning the correlation between the generated numbers. In $4$ out of $8$ stages, $P_{\rm ML}$ surpasses $P_{\rm g}$ in guessing the next random bits by more than $2$ standard deviations (see Fig.~\ref{fig:QRNG:performance}(a)). In the second scenario, when both quantum and electrical noises are considered, the deep learning model shows a considerable decrease in its ability in predicting incoming random numbers. With the presence of quantum noise, only $1$ out of $8$ stages can be predicted better by the RCNN model (Fig.~\ref{fig:QRNG:performance}(b)). Comparing Fig.~\mbox{\ref{fig:QRNG:performance}}(a) and \mbox{\ref{fig:QRNG:performance}}(b), we noticed several features. First, the absolute values of  $P_{\rm ML}$ for Fig.~\mbox{\ref{fig:QRNG:performance}}(a) are $6$--$8$ times higher than those of Fig.~\mbox{\ref{fig:QRNG:performance}}(b). This is due to the fact that the measurement signal is on average around $10$ times stronger than the electronic noise. While the trend of the histograms are similar for Fig.~\mbox{\ref{fig:QRNG:performance}}(a) and (b), we note that the difference and sum for Fig.~\mbox{\ref{fig:QRNG:performance}}(b) are quite different, in contrary to those of Fig.~\mbox{\ref{fig:QRNG:performance}}(a). Since there is no signal present for the latter case, the difference and sum of the two uncorrelated noise in theory gives similar outcomes. For Fig.~\mbox{\ref{fig:QRNG:performance}}(b), the difference and sum probe different quantities, as we shall explain in the next section.

\subsection{Classical entropy}
\begin{figure*}
\centering
\includegraphics[width=17cm]{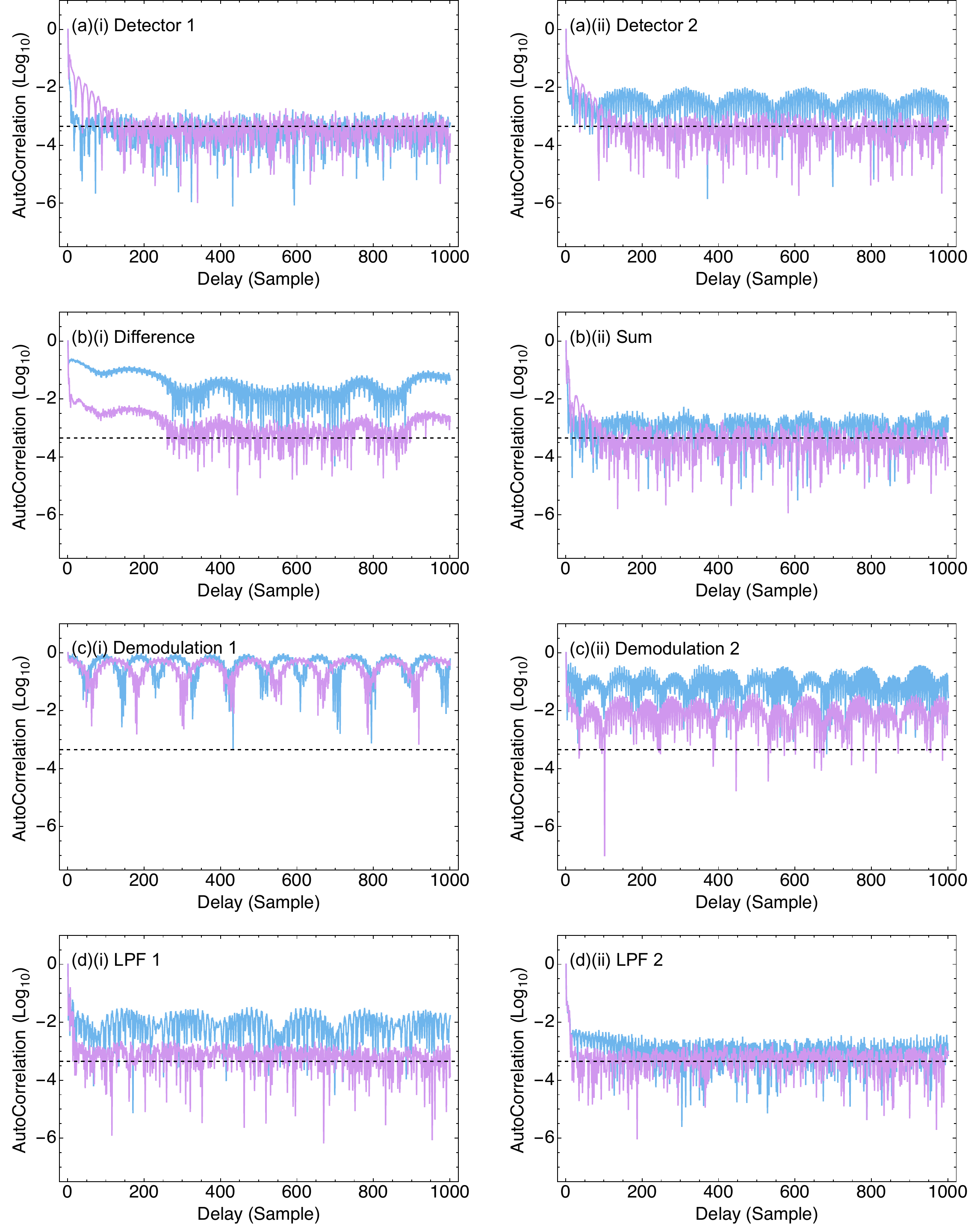}
\caption{Autocorrelation of datasets in scenario 1 (blue), where only electrical noise is present and scenario 2 (pink), where both electrical and quantum noise are present. Dashed lines show the theoretical standard deviation of truly random 5 million samples.}
\label{fig:QAuto}
\end{figure*}

\begin{figure*}
\centering
\includegraphics[width=17cm]{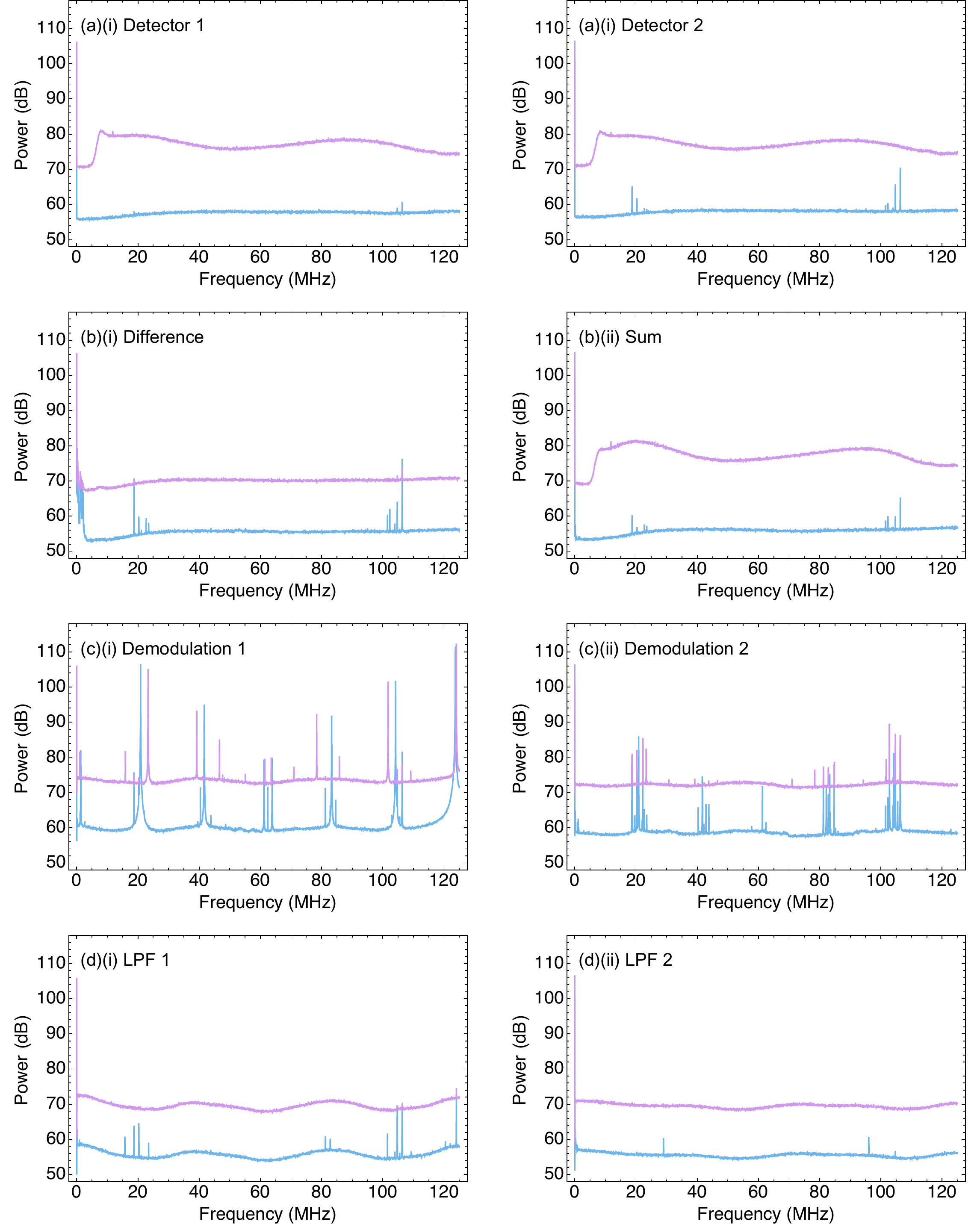}
\caption{Power Spectrum of datasets in scenario 1 (blue), where only electrical noise is present and scenario 2 (pink), where both electrical and quantum noise are present.}
\label{fig:QPow}
\end{figure*}

In order to unravel the underlying reasons behind the advantage offered by the machine learning algorithm, we plot the probability distribution, the autocorrelation and the power spectral density (PSD) of the dataset used for training for both scenarios in Figs.~\ref{fig:Qdistribution}-\ref{fig:QPow}.

Throughout the experiment, the sampling rate is $250$~MSample/s, the Nyquist frequency is half of the sampling rate, i.e.,~$125$~MHz. Within this sampling bandwidth, due to folding, frequencies outside of the bandwidth exist as well. These aliases can compromise the unpredictability of a sequence from the machine learning point of view, as we shall demonstrate below. 

We first look at the first scenario. In stage (a), without the laser light, the signals detected are essentially electrical noise coming from the detectors. Although one of the detectors has persisting nonzero autocorrelation (Fig.~\ref{fig:QAuto}(a)(ii)), the samples from both detectors are normally distributed (Fig.~\ref{fig:Qdistribution}(a)) and have similar frequency response (Fig.~\ref{fig:QPow}(a)). Since the autocorrelation values are close to that obtained from a truly random sample, $P_{\rm ML} \approx P_{\rm g}$. For stage (b), the difference and the sum of the detectors in the homodyne setup are recorded. Interestingly, for the difference of the signals, inherent frequencies in the bandwidth interfered constructively (Fig.~\ref{fig:QPow}(b)(i)), allowing the RCNN algorithm to capture the pattern and predict the outcome better than $P_{\rm g}$. When the signals are demodulated at higher frequencies ($>1$~GHz), the aliases start to contaminate the data collected, as indicated by high autocorrelation values (Fig.~\ref{fig:QAuto}(c)) and prominent pickup frequencies in the detection bandwidth (Fig.~\ref{fig:QPow}(c)). As one would expect, these features give the deep learning model an advantage in determining the next bits. For example, in stage (c)(i), $P_{\rm ML}$ is more than six times higher than $P_{\rm g}$. The application of a low pass filter at the Nyquist frequency, which functions as an anti-aliasing filter in the next stage, mitigates this issue. As a result, both guessing probabilities $P_{\rm g}$ and $P_{\rm ML}$ are comparable to each other.

\subsection{Classical and quantum entropy}
\begin{figure*}
\centering
\includegraphics[width=17cm]{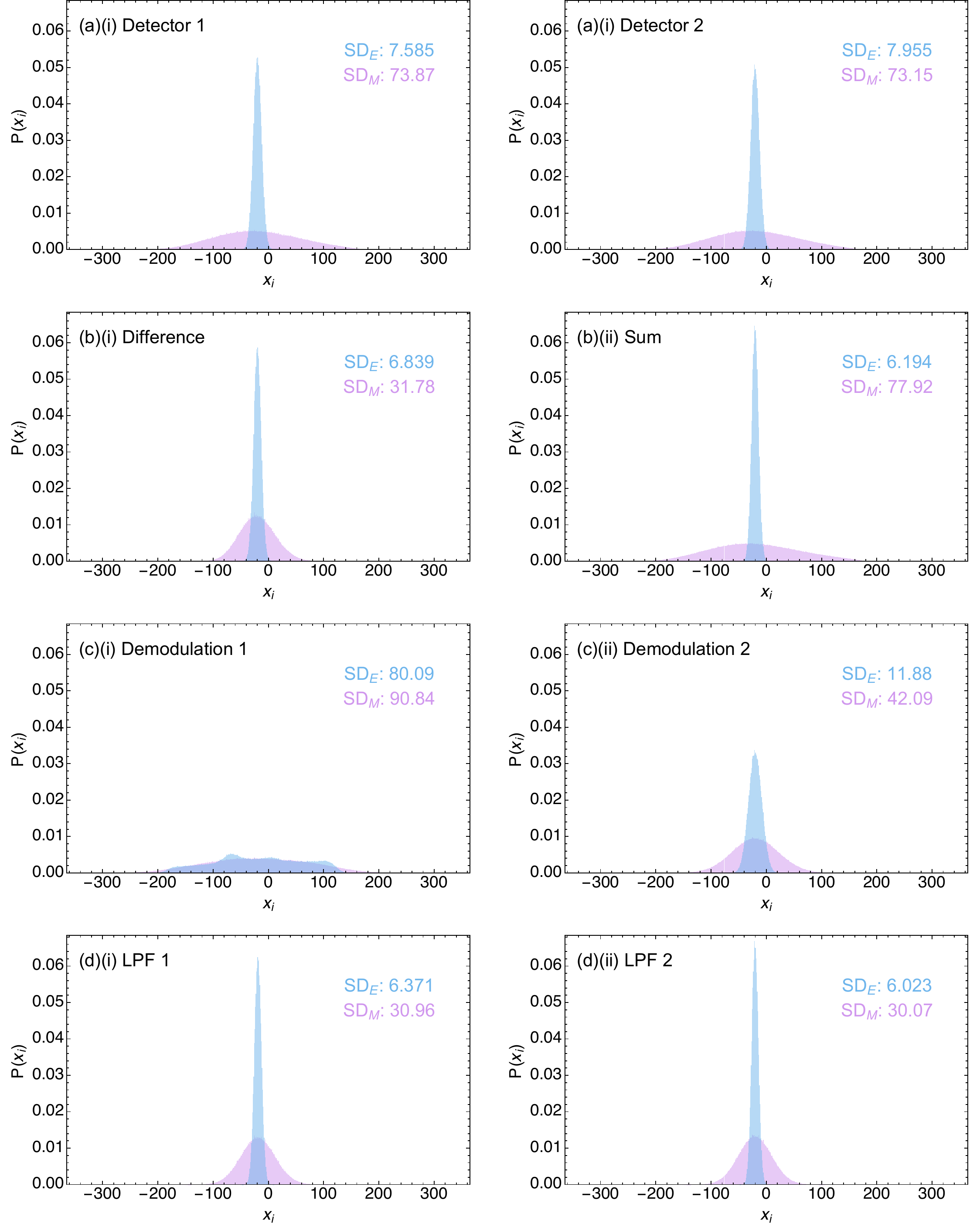}
\caption{Distribution of datasets in scenario 1 (blue), where only electrical noise is present and scenario 2 (pink), where both the electrical and quantum noise are present. ${\rm SD}_{M(E)}$ is the standard deviation of the measured (electronic) signal.}
\label{fig:Qdistribution}
\end{figure*}

For the second scenario, the measured signal is the convolution of the quantum and the electrical signals. As such, the guessing probabilities is lower than the first scenario Fig.~\ref{fig:Qdistribution}. In stage (a), the photocurrent is proportional to the number of photons detected. Since the average intensity of the field is much larger than the fluctuation of the field, the detected photocurrent of each detector in the homodyning can be expressed as~\cite{Lam1998App}
\begin{equation}
\hat{i}(\omega)\propto  \frac{|\alpha|^2}{2} +\frac{\alpha}{\sqrt{2}} \delta\hat{X}(\omega)
\end{equation}
where $|\alpha|^2$ is the intensity of the local oscillator and $\delta\hat{X}(\omega)$ is the amplitude quadrature of the laser field fluctuation at frequency $\omega$. In practice, this photocurrent signal contains electrical noise, which is superposed into the measured probability distribution. Even though there exist oscillations in the autocorrelation, the autocorrelation values eventually converge to that of a uniform distribution. In stage (b), the sum and difference of the homodyne detection are recorded, resulting in the photocurrents with variances given by:
\begin{align}
\Delta^2 i_{\rm Sum}(\omega)& \propto |\alpha|^2 \langle (\delta\hat{X}(\omega))^2)\rangle, \\
\Delta^2 i_{\rm Diff}(\omega)&\propto |\alpha|^2 \langle (\delta\hat{X}_{\rm vac}(\omega))^2)\rangle
\end{align}
As such, the sum and the difference of the photocurrents measure the amplitude quadrature of the local oscillator $\delta \hat{X}(\omega)$ and the vacuum field $\delta \hat{X}_{\rm vac}(\omega)$, respectively. The latter is what we are interested in harnessing to generate random numbers. We remark that the variance of these distributions are quite different in the sidebands of $0-125$ MHz (see Fig.~\ref{fig:Qdistribution}(b)), implying that the sum of the photocurrents is not shot-noise limited within the given bandwidth. For both stages (a) and (b), since the quantum signal is at least 10 dB stronger than the electronic noise floor, and is of the same order of magnitude with most of the technical frequencies (see Figs.~\ref{fig:QPow}(a) \& (b)), no pattern is picked up and $P_{\rm ML}$ is comparable to $P_{\rm g}$. 

For the combine quantum-classical signal scenario, similar to the reason mentioned in scenario 1, the harmonics from sources such as mobile and radio frequencies outside the bandwidth are present, and are stronger than the quantum signal, as shown in Fig.~\ref{fig:QAuto}(c). The autocorrelation values are periodic and finite throughout the sequence as well (Fig.~\ref{fig:QPow}(c)). As a result, $P_{\rm ML} > P_{\rm g}$ for stage (c)(i) and $P_{\rm ML} \approx P_{\rm g}$ for stage (c)(ii). Finally, in stage (d), where LPF is utilized, the autocorrelation is reduced drastically and most spurious frequencies in the sampling bandwidth are eliminated (see Figs.~\ref{fig:QAuto}(d) \& \ref{fig:QPow}(d)). Finite autocorrelation over short delays of the sequence is due to non-ideal filtering~\cite{shen2010practical} and harmonics from sampling. Consequently, $P_{\rm g}$ is always larger than $P_{\rm ML}$ since there is not much context to be learned.

Meanwhile, it is also interesting to examine the effect of additional experimental information on the performance of the ML predictor. For example, we evaluate the correlation between the data of demodulation stages, (c)(i) and (ii) of the QRNG and find non-negligible correlation value of $0.01$. The standard deviation of truly random $N=5 \times 10^6$ samples is given by $1/\sqrt{N} \approx 4.5\times 10^{-4}$. Hence we can expect an increase in $P_{\rm ML}$ with extra information from another frequency band. On the other hand, for the stages after LPFs, (d)(i) and (ii), the correlation between the data is $1.1\times 10^{-4}$, which is consistent with the
correlation between two independent random variables with $5$ million samples. In this case, this uncorrelated additional side information does not help with $P_{\rm ML}$.

\subsection{Entropy source with uniform output}
\label{sec:en_source}
Lastly, we also tested the prediction methods upon the entropy sources with a uniformly distributed output. First, we analyze the final output of our CV-QRNG. Since the raw samples from the homodyne detection is a non-uniform distribution consisting of both the classical noise and the quantum signal (see Sec.~\mbox{\ref{sec:qrng_blk}}), post-processing is necessary for the output of the QRNG to be independent of the entropy of classical origin. To this end, the measurement outcome $M$ is hashed based on the extractable randomness as quantified by min-entropy conditioned upon side information $E$, i.e.,~$H_{\rm min} (M|E)$, the conditional version of Eq.~\ref{eq:Hmindef}. This conditional min-entropy is optimized for a given signal-to-noise ratio, which is given $10\log_{10}(\textrm{SD}^2_M/\textrm{SD}^2_E)$ dB. Here, ${\rm SD}_{M(E)}$ is the standard deviation of the measured (electronic) signal. In our implementation, the Advanced Encryption Standard (AES)~\mbox{\cite{pub2001197}}, a cryptographic hashing algorithm, is used to obtain an almost perfectly uniform output by extracting this secure randomness. Only half of the bits corresponding to the conditional min-entropy is considered secure to ensure the privacy of the random sequence. The full description of the post-processing stage of the QRNG can be found in \cite{haw2015maximization}.

For the evaluation of the deep learning model, we prepare $250$ million of the hashed data in $8$-bit format. We repeat the same RCNN model as described in Sec.~\ref{subsec:DL}. Specifically, we used the first $125$ million $8$-bit numbers to train our RCNN model; the remaining $125$ million numbers were used for testing. The guessing probability $P_{\rm g}$ of the testing sample is equal to the expected probability of a uniform $8$-bit distribution, $1/2^8\approx 0.39\%$. We find that the chance for RCNN to predict the next bit is the same as $P_{\rm g}$, which is as good as making a random guessing. This indicates successful implementation of the QRNG in generating truly random sequences.

\begin{figure}[t]
\centering
	\includegraphics[width=1\columnwidth]{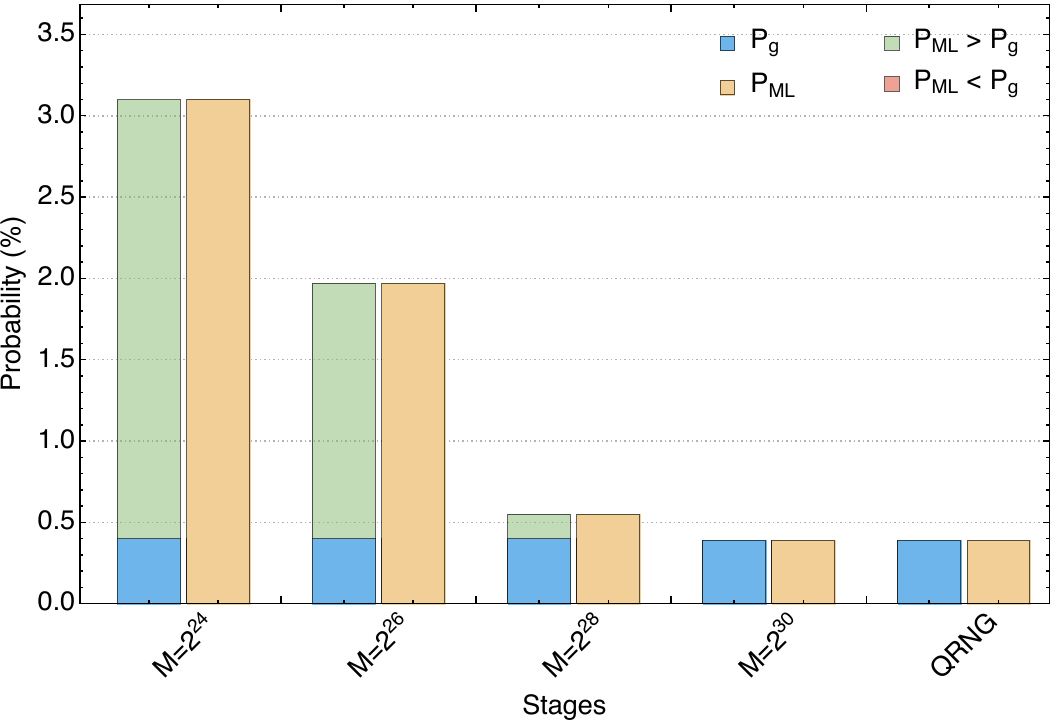}
\caption{Prediction performance of the deep learning model over the uniformly distributed output from QRNG and CRNGs. ML gains advantage over random guessing when the period of the CRNG is less than or comparable to the training data size.}
\label{fig:UD:performance}
\end{figure}

We also verify RCNN model's capability by using it in an attempt to crack a pseudo-random number generator. We choose congruential random number generator (CRNG) as the target. Algorithm for generating numbers used in CRNG is described as below \cite{tezuka1995LCG},

$$X_{n}=(a X_{n-1} + c) (\textrm{mod}\mathcal{M}),$$
$$u_n=\frac{X_n}{\mathcal{M}},$$

where $a$,$c$, and $\mathcal{M}$ are integers. The output $u_n$ is a random number sequence in $[0,1)$. We collected $250$ million  $8$-bit numbers  generated by CRNG with $a=1103515245$, $c=12345$ and $\mathcal{M} \in (2^{24}, 2^{26}, 2^{28}, 2^{30})$. We apply the same approach in splitting data for training and testing. Because the choice of $a$, $c$ and $\mathcal{M}$ satisfies three conditions: (1) $\mathcal{M}$ and $c$ are relatively prime, (2) $a-1$ is divisible by all prime factors of $\mathcal{M}$, and (3) $a-1$ is divisible by $4$ if $\mathcal{M}$ is divisible by $4$, $\mathcal{M}$ is equal to the period of the sequence generated by the CRNG~\mbox{\cite{tezuka1995LCG}}. We are interested in how different periods have an impact on the ML. The RCNN model achieves $3.13 \pm 0.03\%$, $1.97 \pm 0.01\%$, $0.55 \pm 0.01\%$, $0.39 \pm 0.01\%$ accuracy in predicting next generated $8$-bit number given $10$ previous numbers when $\mathcal{M}$ is $2^{24}$, $2^{26}$, $2^{28}$, $2^{30}$, respectively (Fig.\mbox{~\ref{fig:UD:performance}}). Given the same length of training random number sequence, $P_{\textrm{ML}}$ decreases when $\mathcal{M}$ increases, i.e., the period increases. Note that with $\mathcal{M}=2^{28}$ which is two times larger than the training size, ML can still have $P_{\textrm{ML}}$ better than $P_{\textrm{g}}$ by more than 15 standard deviations. With larger $\mathcal{M}$, i.e., $\mathcal{M}=2^{30}$ or higher, $P_{\textrm{ML}} \approx P_{\textrm{g}}$.

To elucidate the performance of RCNN technique further, we run the NIST statistical test suite (STS) \cite{rukhin2001cryto} upon the test sets from both the QRNG and CRNGs. This STS consists of $15$ empirical tests of randomness aimed at the testing of both hardware or software based RNG. The result of each test is a $p$-value that represents the probability that the chosen test statistics will assume values that are equal to or worse than the observed test statistics. The result is shown in Table~\mbox{\ref{Tab:Result:nist}}. While both approaches aim at detecting deviations from randomness, they are operationally different. For the statistical test, the uniformity of a sequence is examined through various algorithms. Our ML, on the other hand, targets at the unpredictability aspect of an RNG by learning the previously generated sequence to guess the next output bit. The output from the QRNG passes the NIST test successfully, in agreement with the fact that ML found no pattern in the data. For the CRNGs, the number of passes tests increases with the period $\mathcal{M}$. When the period is long enough compared to the size of the test set, i.e. for $\mathcal{M}=2^{28}$ and $2^{30}$, the CRNG passes the NIST STS successfully. Comparing with the result from Fig.~\mbox{\ref{fig:UD:performance}}, we note that for $\mathcal{M}=2^{28}$, even though the NIST test is positive, ML gives better prediction than a random guessing.
\begin{table*}[hbt]
\centering
\caption{Results of NIST statistical tests~\mbox{\cite{rukhin2001cryto}} for CRNG and QRNG. For each test set, the data size is $1$~Gbits ($1000$ sequences with each sequence around $1$~Mbits). To pass the test, the uniformity of the $p$-values ($P$-value) should be larger than 0.0001 and the proportion of the sequences with $p$-values larger than the significant level $\alpha=0.01$ must be in the range of $0.99\pm0.0094392$. For the random-excursions (variant), the range is $0.99\pm0.0119303$. When the evaluation produces multiple $P$-values (marked with $^\dagger$), the worst case is shown.
} \label{Tab:Result:nist}
\begin{tabular}{lccccccc}
\toprule
\toprule
  &  & \multicolumn{2}{c}{CRNG}  & & \multicolumn{3}{c}{QRNG} \\
\cmidrule(r){2-5}  \cmidrule(r){6-8}

      Evaluation & $\mathcal{M}=2^{24}$ & $\mathcal{M}=2^{26}$ & $\mathcal{M}=2^{28}$ & $\mathcal{M}=2^{30}$ & $P$-value& Proportion & Result  \\

\midrule
    
  Frequency &\emph{failure}  &success &success& success & $0.558502$ & $0.9840$ & success \\
  Block-frequency   & success   &success&success & success & $0.492436$ & $0.9950$ & success \\
  Cumulative sums $^\dagger$  &\emph{failure}    &success&success & success & $0.360287$ & $0.9880$ & success \\
  Runs   & success  &success&success & success & $0.420827$ & $0.9860$ & success \\
  Longest run & success  &success &success & success & $0.680755$ & $0.9940$ & success \\
  Rank  & success  &success& success & success  & $0.382115$ & $0.9890$ &success \\
  FFT  & \emph{failure}  &success &success & success  & $0.083018$ & $0.9890$ & success \\
  Non-overlapping template $^\dagger$ & \emph{failure}  &success &success & success & $0.016602$ & $0.9860$ & success \\
  Overlapping template  & success  &success  &success & success&  $0.639202$  & $0.9900$ & success \\
  Universal  & success  &success&success & success & $0.595549$ & $0.9870$ & success \\
  Approximate entropy    &\emph{failure}  &success  &success& success & $0.739918$ & $0.9910$ & success \\
  Random-excursions $^\dagger$ & success &success &success & success & $0.284657$ & $0.9904$ & success \\
  Random-excursions variant $^\dagger$ &success & success &success & success & $0.007210$ & $0.9904$ & success \\
  Serial $^\dagger$  &\emph{failure}  &\emph{failure}  &success & success & $0.339271$ &  $0.9930$ & success \\
  Linear-complexity  &success &success &success & success & $0.181557$ & $0.9910$ & success \\
  \midrule
  Total successful tests & $9/15$ & $14/15$ & $15/15$ & $15/15$ & - & - & $15/15$ \\
\bottomrule
\bottomrule
\label{tbl:QRNG:test}

\end{tabular}

\end{table*}

\section{Discussion and Conclusion}
To summarize, we have applied the machine learning algorithm to examine potential patterns in the raw entropies from a QRNG setup. In particular, we analyze the data from different stages of the continuous variable QRNG in \cite{haw2015maximization}. 
While the classical noise themselves have large entropy, they are more prone to interference from the deterministic source as shown by our deep learning results. For the entropy source consisting of quantum and classical noise, although the quantum effect is the dominant entropy source, due care must be taken to ensure that the effect of the hidden correlations such as memory effects and biases are minimized~\cite{herrero2017quantum}. To further bound the effect of the random classical noise, which ultimately could be deterministic, quantification of entropy independent of the classical noise or the presence of the eavesdropper should be used~\cite{haw2015maximization,marangon2017source}.

By applying the machine learning technique upon the final output data, we verified that upon proper entropy quantification and post-processing, the randomness of the uniform output is guaranteed. This is contrasted with a uniform output from a pseudo RNG based on linear congruential generator. A comparison between our ML approach and traditional statistical test is in order. Although the ML can be computationally more demanding compared to randomness statistical test such as NIST and Dieharder battery test, we believe the relevance of ML is twofold: Firstly, since our ML is based on neural network, it could potentially reveal aspects overlooked by the mathematical algorithm. As shown in Sec.~\mbox{\ref{sec:en_source}}, our ML can predict better than random guessing even when the sequence in examination passed all the NIST test. Secondly, the scope of a statistical test usually is limited to evaluating the randomness of a uniform distribution. In our work, we show that the combination with the guessing probability, $P_{\rm g}$ allows one to examine the unpredictability of non-uniform distribution as well. By comparing with the guessing probability $P_{\rm g}$, our scheme is able to indicate potential hidden correlation in both uniform and non-uniform distribution.

Future applications of our method bridge the gap between the theoretical and experimental description of the device. For example, we can examine the ML with different experimental parameters, such as digitization level, SNR ratio, ADC range and sampling speed. By studying the extent of the influences of the deterministic classical noise under these different settings, one can identify an optimal operating regime for the QRNG without compromising the desired figure of merit. For post-processing in RNG, ML prediction technique can be utilized to study the effectiveness of simple algorithm in unbiasing, such as Von Neumann extractor and XOR operation. One could potentially analyze the robustness of a randomness extractor over different input parameters with ML. For example, for a Toeplitz hashing algorithm, it is interesting to find out how different parameters such as the input length and the security parameter change the predictive power of ML. Since our scheme is RNG-agnostic, it can serve as a versatile tool in evaluating the unpredictability of any hardware random number generator.



\section{Acknowledgement}
This research was supported by Sydney Informatics Hub, funded by the University of Sydney. OK acknowledges the support of Australian Research Council (ARC) under grant DP140103448. NDT acknowledges support of the Commonwealth Scientific and Industrial Research Organisation (CSIRO) under grant PN~50041400. JYH, SMA and PKL acknowledge the support of Australian Research Council (ARC)
under the Centre of Excellence for Quantum Computation and
Communication Technology (project number CE110001027).


%





\ifCLASSOPTIONcaptionsoff
  \newpage
\fi



%
\bibliographystyle{IEEEtran}
\bibliography{MyCollection_abbr}

%









\vspace*{-1.2cm}
\begin{IEEEbiography}[{\includegraphics[width=1in,height=1.25in,clip,keepaspectratio]{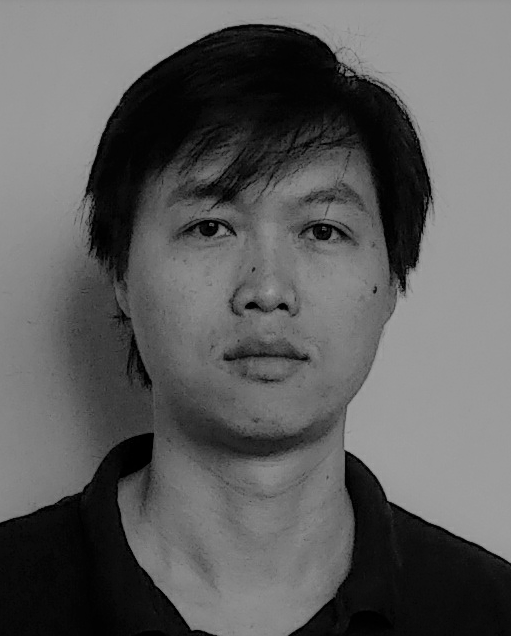}}]{Nhan Duy Truong} is a PhD candidate at Royal Melbourne Institute of Technology, Australia. He has completed his B.Eng. in Electronics and Telecommunications with Gold Medal for Top Graduate from Ho Chi Minh City University of Technology, Vietnam in 2010. He received his M.Eng. in Electronics and Computer Engineering in 2013 at Royal Melbourne Institute of Technology. His research interests include neural networks, medical devices, deep learning and its hardware implementation.
\end{IEEEbiography}

\vspace*{-1.2cm}
\begin{IEEEbiography}[{\includegraphics[width=1in,height=1.25in,clip,keepaspectratio]{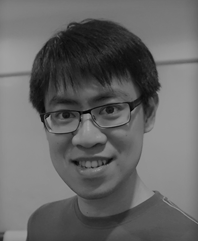}}]{Jing Yan Haw}
received his Bachelor degree in Physics with First Class Honours from the National University of Singapore. He participated in an intensive internship program as a quantum optics theorist at the University of Tokyo before joining the Centre for Quantum Technologies at the National University of Singapore in 2011 as a research assistant. Haw completed his PhD dissertation on continuous variable quantum random number generation and probabilistic linear amplification and worked as a research associate at the Australian National University in 2017.
\end{IEEEbiography}

\vspace*{-1.2cm}
\begin{IEEEbiography}[{\includegraphics[width=1in,height=1.25in,clip,keepaspectratio]{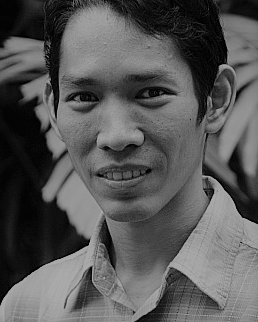}}]{Syed Muhamad Assad} completed his BSc with a double major in Physics and Computational Science at the National University of Singapore. He worked at the National University of Singapore as a teaching assistant and at the Centre for Quantum Technologies as a research assistant. He obtained joint PhD between the National University of Singapore and the Australian National University in 2011. The main subjects of his PhD were the realisation of Harmonic Entanglement between a light beam and its second harmonic, and theoretical proofs of security for Quantum Key Distribution protocols. He has been working in the quantum optics group at the Australian National University since January 2011. He is a member of the Centre of Excellence for Quantum Computation and Communication Technology. He is currently leading the secure communications team in the quantum optics group at the Australian National University.
\end{IEEEbiography}

\vspace*{-1.2cm}
\begin{IEEEbiography}[{\includegraphics[width=1in,height=1.25in,clip,keepaspectratio]{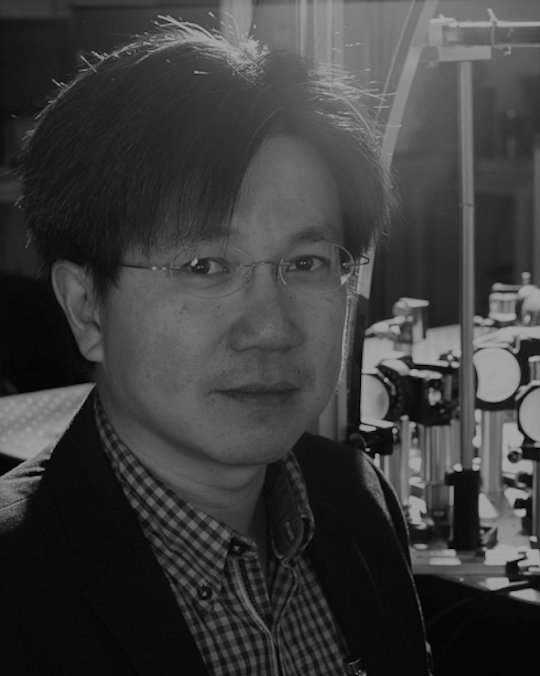}}]{Ping Koy Lam} completed his BSc with a double major in Maths and Physics from the University of Auckland. He worked as a process engineer for Sony (audio electronics) and Hewlett-Packard  (semiconductor LED) for 3 years prior to his post-graduate studies at the ANU where he obtained an MSc in theoretical physics, and a PhD in experimental physics. His dissertation was on the generation of squeezed light, quantum electro-optic control, and quantum teleportation. He was awarded the Australian Institute of Physics Bragg Medal and the Australian National University Crawford Prize for his PhD dissertation in 1999.

Apart from his position at the Australian National University, Lam has worked at the Erlangen-Nürnberg Universität and Paris University as Alexander von Humboldt Fellow and CNRS Visiting Professor, respectively. He was also an adjunct Professor for Tianjin University from 2013 to 2015. He was awarded the British Council Eureka Prize for inspiring science with his research and outreach activities, and the University of New South Wales Eureka Prize for innovative research for his work in quantum communication. He also co-founded QuintessenceLabs – the first Australian company to commercialize quantum communication technology. He is currently the ANU node director for the Centre of Excellence for Quantum Computation and Communication Technology and the Australian Research Council Laureate Fellow. Lam has published around 260 articles with more than 40 papers in Physical Review Letters, Science, and the Nature research journal suite. 
\end{IEEEbiography}

\vspace*{-1.2cm}
\begin{IEEEbiography}[{\includegraphics[width=1in,height=1.25in,clip,keepaspectratio]{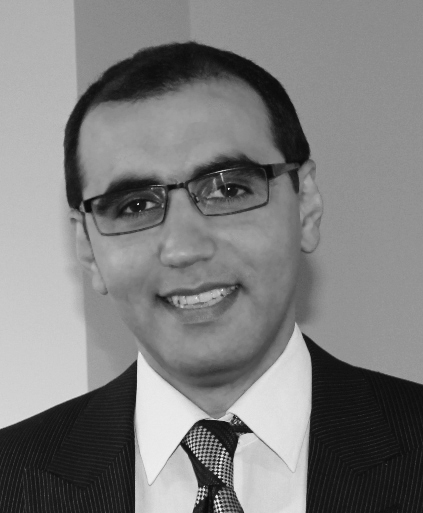}}]{Omid Kavehei} (S'05-M'12-SM'16) is a Senior Lecturer at The University of Sydney, Australia. Prior to this role he was with Centre for Neural Engineering at The University of Melbourne as a Research Fellow in Microelectronics, where he worked on an Australian engineering flagship project, the Bionic Eye. He was also a tenured Lecturer with Royal Melbourne Institute of Technology, where he is holding an Honorary position. He received his PhD degree in Electronic Engineering from The University of Adelaide in 2012 with Faculty Dean's Commendation for Academic Excellence and 2013 Post-Graduate University Alumni Medal. His research interests include custom-design hardware-accelerated machine intelligence for medical, health-care and cyber-security applications, emerging solid-state memory devices, implantable electronics, cyber-physical systems and  security, and novel computational paradigms with nanotechnology. He was an Executive Member of the South Australia IEEE student chapter and the recipient of several awards and fellowships including an Endeavour Research Fellowship in 2017 and South Australian Young Nanotechnology Ambassador Award in 2011.
\end{IEEEbiography}

\end{document}